\documentclass{article} % For LaTeX2e
\usepackage[preprint]{colm2026_conference}

\usepackage{microtype}
\usepackage{hyperref}
\usepackage{url}
\usepackage{booktabs}

\usepackage{lineno}

\definecolor{darkblue}{rgb}{0, 0, 0.5}
\hypersetup{colorlinks=true, citecolor=darkblue, linkcolor=darkblue, urlcolor=darkblue}

\title{Disentangling Language Modeling and Boundaries}

\author{Mykola Haltiuk\\
Faculty of Computer Science\\
AGH University of Krakow\\
Krakow, Poland \\
\texttt{mhaltiuk@agh.edu.pl} \\
}

\begin{document}

\ifcolmsubmission
\linenumbers
\fi

\maketitle

\begin{abstract}
Byte-level language models are usually argued for on the grounds of robustness, multilingual fairness, and character-level skills. We point to a different, structural advantage: because they read and write bytes, any two of them share an output space, so knowledge transfer between them is exact and independent of how either was originally tokenized. We hypothesize that the two distributions a byte-level model produces, one over the next byte, one over where its patch boundaries fall, can be disentangled and changed almost independently. A model could absorb a teacher's capability while keeping its own boundaries, or change how it places those boundaries while keeping its capabilities. We lay out the two experiments that would settle the hypothesis, alongside preliminary measurements of the properties they rest on. We argue that the community should move toward a byte-level interface as a shared standard: if the hypothesis holds, then once byte-level models are the norm, transferring capabilities and reshaping boundaries between them become cheap and routine, free of the per-model tokenizer that blocks them today.
\end{abstract}

\section{Position}

Most of the language modeling research, especially in non-English settings, plays catch-up with the newest capabilities of frontier models. One of the cheapest ways to pass on a skill would be logit distillation~\citep{hinton2015distilling}, yet, in practice, differing tokenizers often limit the student's ability to mimic the output distribution of the teacher, since the two distributions are defined over different vocabularies. The problem bites hardest where distillation would help most. For mid- and low-resource language model adaptation, strong teachers exist, but instruction and reasoning data are scarce, so practitioners either keep a strong model's tokenizer~\citep{MamayLMv1,yukhymenko-etal-2026-recovered}, which lets them distill new capabilities from a same-tokenizer teacher, and accept its poor compression on the target language, or change it through tokenizer transfer~\citep{dobler2026token, haltiuk2026modelawaretokenizertransfer}, but then suffer the costs of extensive synthetic data generation~\citep{paniv2026dataefficient} to instill new capabilities. Cross-tokenizer distillation~\citep{boizard2025towards, NEURIPS2025_720f9f5d} works across vocabularies, but only approximately, not with the exact signal a shared vocabulary gives.

We argue that the tokenizer should not be part of the interface at all. A language model should read and write bytes, regardless of how it processes them internally. ByT5~\citep{xue-etal-2022-byt5} processes every byte without any aggregation; BLT~\citep{pagnoni-etal-2025-byte} groups bytes into patches by next-byte entropy; H-Net~\citep{hwang2025dynamic} learns where to split end to end. Bolmo~\citep{minixhofer2025bolmo} byteifies an existing subword model, distilling its original tokenizer into an internal boundary predictor, effectively separating the language modeling from boundary prediction. Once models share a byte interface, the obstacle behind every cross-tokenizer method disappears. Both models emit distributions over the same 256 byte values, and exact distillation on next-byte predictions becomes straightforward. Language modeling, though, is only one of the two distributions a byte-level model produces, while the other places its patch boundaries. If the two can be disentangled, a model could take on any teacher's skill while keeping its own way of reading text, and change that way without forgetting what it knows.

\section{Preliminaries}

Replacing a model's embeddings with a byte interface is known not to erase its capabilities, since they live in the deeper layers rather than the embeddings~\citep{minixhofer2025bolmo}, and our initial results confirm embedding resettability holds on Ukrainian models too.

Then, to even talk about disentangling the language modeling and boundary prediction distributions, we first need to be able to measure both. While language modeling is straightforward, we treat boundary prediction as a per-byte binary classification problem, where each byte position is either a boundary or not, and define the boundary divergence between two models as $1 - F_1$. We have already conducted some initial measurements across different domains and model families, and found promising, high-divergence settings for our main experiments, most clearly observed between SentencePiece- and BPE-derived models. We have also measured how far the Bolmo model's predicted boundaries diverge from the OLMo-family tokenizer it was byteified from, and found that they barely differ at all, which implies a Bolmo-style model can learn a given boundary distribution relatively easily.

\section{Disentangling language modeling from boundary prediction}

Our central hypothesis is that these two distributions can be disentangled, so that either one can be changed without meaningfully disturbing the other. Bolmo is a convenient architecture for testing this: its output is a joint distribution $p(b, m)$ over a byte $b$ and a boundary marker $m$ (every byte appears in the vocabulary twice, plain and patch-ending). The language modeling and boundary distributions are its marginals $\sum_m p(b, m)$ and $\sum_b p(b, m)$, which we can manipulate separately even though the model emits them together.

Our first experiment will distill one Bolmo model into another over the language modeling distribution alone, leaving each model's boundaries untouched. Since both models write to the same byte alphabet, the transfer is exact rather than approximate, and the question we aim to answer is whether a capability moves from teacher to student while the student keeps placing boundaries its own way, which we will check by measuring the boundary divergence before and after. This is close in spirit to recent work that distills across tokenizers through a byte-level interface~\citep{singh2026crosstokenizer}, but where that work bolts a temporary byte head onto a subword student and removes it afterwards, here the byte interface is simply what both models already are, so any byteified model can teach or learn from any other without building a bridge for the pair.

Our second experiment turns the same idea around: instead of changing what the model knows while keeping its boundaries, we will try to change them while keeping what it knows -- in effect a tokenizer transfer for byte-level models, where the boundary predictor plays the role the tokenizer plays in a subword model. If this works, it would give us a far more direct way to control a model's compression rate than the merging heuristics explored in Bolmo. We would self-distill the language modeling distribution to keep it fixed, while retraining both boundary predictors: the auxiliary one and the head that actually drives patching at decode, so that the two stay in agreement. We expect this to be harder than it sounds, since the boundaries decide how bytes are pooled into patches and so reshape the very input the backbone transformer sees, the same backbone that BLT, H-Net and Bolmo all rely on.

If both directions hold, the consequences reach past any single model. Byteifying a model is relatively cheap, Bolmo does it for under a percent of pretraining cost, so a byte-to-byte route could first become simply a better way to do the cross-tokenizer transfer people already attempt, and over time a reason to drop the per-model tokenizer altogether in favor of a shared interface. We are aware that bytes are not a perfect choice here. UTF-8 is biased toward Latin scripts, spending more bytes per character on exactly the languages that motivate this work, even if patching hides much of that cost inside the cheap local encoder. We therefore see the deeper question not as whether to use bytes, but as what the right common alphabet for language models should be, and we encourage the community to take it up.

% \section*{Author Contributions}

% \section*{Acknowledgments}
% Use unnumbered first level headings for the acknowledgments. All

% \section*{Ethics Statement}
% Authors can add an optional ethics statement to the paper. 
% For papers that touch on ethical issues, this section will be evaluated as part of the review process. The ethics statement should come at the end of the paper. It does not count toward the page limit, but should not be more than 1 page. 

\bibliography{colm2026_conference}
\bibliographystyle{colm2026_conference}

% \appendix
% \section{Appendix}

\end{document}